# Variational Autoencoding Molecular Graphs with Denoising Diffusion Probabilistic Model

Daiki Koge
*Division of Information Science, Graduate School of Science and Technology, Nara Institute of Science and Technology,*
Takayama, Ikoma, Nara, Japan
koge.daiki.ju9@is.naist.jp

Naoaki Ono
*Division of Information Science, Graduate School of Science and Technology, Nara Institute of Science and Technology,*
Takayama, Ikoma, Nara, Japan
*Data Science Center, Graduate School of Science and Technology, Nara Institute of Science and Technology,*
Takayama, Ikoma, Nara, Japan

Shigehiko Kanaya
*Division of Information Science, Graduate School of Science and Technology, Nara Institute of Science and Technology,*
Takayama, Ikoma, Nara, Japan
*Data Science Center, Graduate School of Science and Technology, Nara Institute of Science and Technology,*
Takayama, Ikoma, Nara, Japan

*Abstract*—In data-driven drug discovery, designing molecular descriptors is a very important task. Deep generative models such as variational autoencoders (VAEs) offer a potential solution by designing descriptors as probabilistic latent vectors derived from molecular structures. These models can be trained on large datasets, which have only molecular structures and applied to transfer learning. Nevertheless, the approximate posterior distribution of the latent vectors of the usual VAE assumes a simple multivariate Gaussian distribution with zero covariance, which may limit the performance of representing the latent features. To overcome this limitation, we propose a novel molecular deep generative model that incorporates a hierarchical structure into the probabilistic latent vectors. We achieve this by a denoising diffusion probabilistic model (DDPM). We demonstrate that our model can design effective molecular latent vectors for molecular property prediction from some experiments by small datasets on physical properties and activity. The results highlight the superior prediction performance and robustness of our model compared to existing approaches.

*Keywords—Molecular descriptors, Variational Autoencoders, Transfer Learning, Denoising Diffusion Probabilistic Model*

## I. Introduction

Accurately predicting the physical and chemical properties of molecules has always been a topic of interest in cheminformatics. Designing molecular descriptors that quantify the molecular structure is an important factor affecting the accuracy of molecular property prediction. Recent graph neural networks [1] are promising approaches to tackle this task, however, the generalization performance of these models is often constrained by the scarcity of available data. Deep generative models, such as variational autoencoders (VAEs) [2], offer a potential solution by designing descriptors as probabilistic latent vectors encoded from molecular structures. These models can be trained on large datasets, such as ZINC [3], which have only molecular structures.

VAEs consist of two networks: an Encoder that encodes a molecular structure into a low-dimensional probabilistic latent vector following a certain probability distribution, and a Decoder that generates a molecular structure from the latent vector (Fig.1). During training, the VAEs are optimized to reconstruct the original molecular graph from the latent vector sampled from the Encoder. Thus, it is possible to extract a latent vector that preserve the information of the entire a molecular structure from the encoder [4][5]. VAEs utilize the variational bayesian method to infer the approximate posterior distribution of the latent vector $z$ using the encoder.

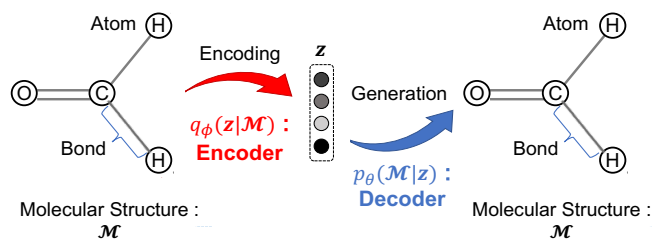

Figure 1. *Overview of the VAEs framework for molecules.*

In the original VAEs, the encoder $q_\phi(z|\mathcal{M})$ infers the approximate posterior as a multivariate gaussian distribution $N(z; u_\phi(\mathcal{M}), diag(\sigma_\phi^2(\mathcal{M})))$ with zero covariance. The simplicity of this approximate posterior distribution for embedding molecular structures may limit the performance in representing the latent features.

In this study, we propose a novel molecular deep generative model that improves the performance of representing the approximate posterior distribution of VAEs by applying a denoising diffusion probabilistic model (DDPM) [6]. Our model encodes a molecular graph into a hierarchical probabilistic latent vector by diffusion process.

## II. Methods

An overview of the proposed model is shown in Fig.2. Here, a molecular graph $\mathcal{G}$ is defined as a set of nodes (atoms): $V$ and a set of edges (bonds): $E$. Our approach is to map a molecular graph $\mathcal{G}$ into $d$-dimensional probabilistic latent vectors $(z_1, z_2, ... z_T)$ using a transformer [7] and model the probability distribution of these latent vectors using denoising diffusion probabilistic model.

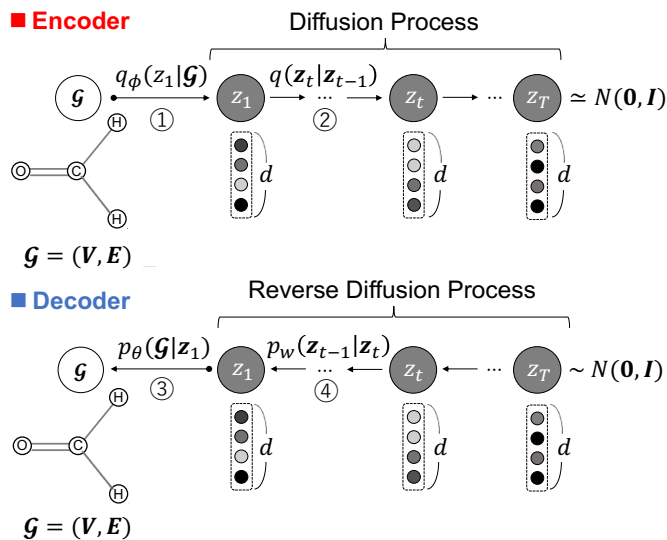

Figure 2. *Overview of our proposed model.*

The approximate posterior distribution of a latent vector at each time $t$ in the encoder scheme represented by ① and ② in Figure 2 is defined as follows.

$$q_\phi(z_1|\mathcal{G}) = N(z_1; \sqrt{1-\beta_1}f_\phi(\mathcal{G}), \beta_1 I), \quad (1)$$
$$q_\phi(z_t|z_{t-1}) = N(z_t; \sqrt{1-\beta_t}z_{t-1}, \beta_t I), \quad (2)$$

where $f_\phi$ is a mapping $\mathcal{G} \to z_0 \in \mathbb{R}^d$ using a transformer. These conditional probabilities represent the Markov property of the probabilistic latent vectors $(z_1, z_2, ... z_T)$ and approximate $q_\phi(z_T|\mathcal{G})$ to the normal distribution $N(z_T; 0, I)$ according to a variance schedule $\beta_1, ..., \beta_T$. The decoding process ③ and ④, which represent decoding from the probabilistic latent vectors to the observed data (molecular graph $\mathcal{G} = (V, E)$), is defined by the following equations.

$$p_\theta(V, E|z_1) = \prod_{v=1}^{V_N}\prod_{k=1}^{K} f_\theta(z_1)_{(v_k)}^{v_k} \cdot \prod_{e=1}^{E_N}\prod_{i=1}^{L} f_\theta(z_1)_{(e_i)}^{e_i}, \quad (3)$$

$$p_w(z_{t-1}|z_t) = N(z_{t-1}; u_w(z_t, t), \sigma_t^2 I), \quad (4)$$

$$u_w(z_t, t) = \frac{1}{\sqrt{\alpha_t}}\left(z_t - \frac{\beta_t}{\sqrt{1-\overline{\alpha_t}}}\epsilon_w(z_t, t)\right), \quad (5)$$

$$\alpha_t \equiv 1 - \beta_t \text{ and } \overline{\alpha_t} \equiv \prod_{s=1}^{t} \alpha_s, \quad (6)$$

where $f_\theta$ is a mapping $z_1 \to (V, E) \in \mathbb{R}^{V_N \times K} \times \mathbb{R}^{E_N \times L}$ using a transformer with softmax activation. And we denote the probability values for each element ($v_k$ and $e_i$) in $V$ and $E$ by $f_\theta(z_1)_{(v_k)}$ and $f_\theta(z_1)_{(e_i)}$, assuming that each element is sampled from a categorical distribution. $\epsilon_w(z_t, t)$ in Eq. 5 is a multi-layer perceptron (MLP) with a skip connection. Our model integrates PIGVAE [5] and DDPM [6].

The objective of our model is evidence lower bound (ELBO) $\mathcal{L}(\theta, \phi, w)$ on the log marginal likelihood $\log p_\theta(\mathcal{G})$:

$$\mathcal{L}(\theta, \phi, w) = \mathbb{E}_{q_\phi(z_1, ... z_T|\mathcal{G})}[\log p_\theta(\mathcal{G}|z_1)]$$
$$- D_{KL}[q_\phi(z_T|\mathcal{G}) \| N(z_T; 0, I)]$$
$$- \mathbb{E}_{t \sim \text{Uniform}(\{1,...T\})}[\|\varepsilon_t - \epsilon_w(z_t, t)\|^2],$$

where $\varepsilon_t$ is a gaussian noise added at time $t$ during the diffusion process.

## III. EXPERIMENTS

We empirically evaluated our model on molecular property prediction tasks for small datasets in the following two steps: 1) training a molecular VAE (PIGVAE) with transformer and our model for ZINC dataset, and 2) fine-tuning trained models for molecular property regression with small datasets. Each dataset was divided into training data and test data for evaluation using hold out method. When fine-tuning the VAE and our model, we defined a loss function that combined the ELBO objective for a molecular graph with the prediction error (Mean Squared Error) obtained from a molecular property regression with an MLP model using a probabilistic latent vector from the encoder as the input. In our model, we used $z_1$ as the input which was obtained in Eq. 1.

TABLE 1. Prediction errors for each dataset and model.

| Models | Small Dataset | | | |
|---|---|---|---|---|
| | BACE1 | CTSD | FreeSolv | Lipophilicity |
| Graph Transformer | 0.5362 | 0.4941 | 1.7611 | 0.6422 |
| PIGVAE [5] | 0.4707 | 0.3543 | 1.7583 | 0.6106 |
| Our Model | 0.4238 | 0.2593 | 1.2074 | 0.5823 |

We summarized mean squared errors of each model in the test data for each dataset in Table 1. BACE1 and CTSD are binding affinity data for target proteins obtained from ExCAPE-DB [8]. FreeSolv and Lipophilicity are physico-chemical property data obtained from MoleculeNet [9]. We also included results from an encoder-only model (Graph Transformer). This model uses the transformer in the PIGVAE and is not pre-trained by ZINC. Out method showed better prediction performance than the other methods.

To visualize the latent space of molecules, we embedded the molecular latent vectors into two-dimensional space using Umap [10]. We trained PIGVAE and our model with QM9 dataset [11] and colored the samples on the 2D latent space (Figure 3). Surprisingly, even though our model didn't use HOMO energy for training, we found that the HOMO energy was smoothed on the latent space compared to the VAE model. This could be the reason for the good prediction of physico-chemical properties such as lipophilicity.

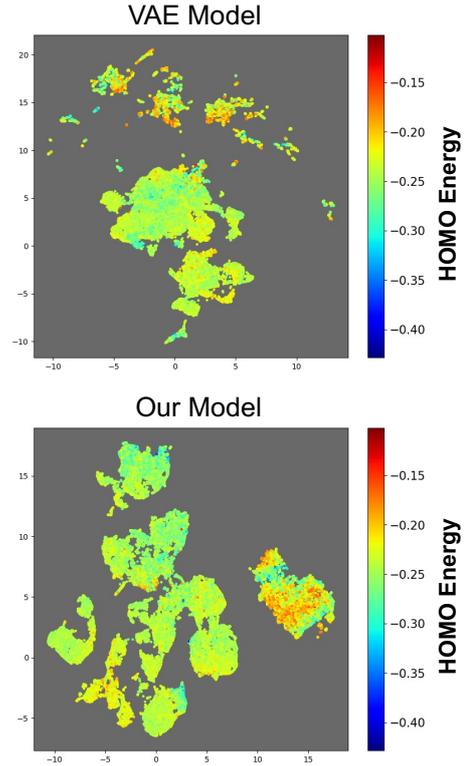

Figure 3 : Visualization of latent space QM9 dataset

## IV. CONCLUSION

In this study, we proposed a novel probabilistic embedding method for molecular graphs. Our model integrates graph transformer for molecular graphs and DDPM, allowing molecular graphs to be encoded in hierarchical probabilistic latent vectors. Consequently, our model was able to encode molecular structures into flexible latent vectors, yielding higher prediction performance than the existing VAE models and a graph transformer.

ACKNOWLEDGMENT

This work was supported by JSPS KAKENHI Grant Number 22J11040.